\documentclass[dvipsnames]{elsarticle}

\usepackage{lineno}
\usepackage[colorlinks]{hyperref}
\modulolinenumbers[5]

\usepackage{subcaption}
\usepackage{tikz}
\usepackage{pgfplots}
\usepackage{xcolor}
\usepackage{colortbl}
\usepackage{amsmath}
\usepackage{amsfonts}
\usepackage{mathtools}
\usepackage{bm}

\usetikzlibrary{arrows.meta}
\usepgfplotslibrary{groupplots}

\newcommand{\C}{\mathcal{C}}
\newcommand{\D}{\mathfrak{D}}
\newcommand{\R}{\mathbb{R}}
\newcommand{\uu}{\mathbf{u}}
\newcommand{\vv}{\mathbf{v}}
\newcommand{\x}{\mathbf{x}}
\newcommand{\XX}{\mathbf{X}}
\newcommand{\X}{\mathcal{X}}
\newcommand{\Y}{\mathcal{Y}}
\newcommand{\B}{\mathcal{B}}
\newcommand{\problem}{\mathcal{P}}
\newcommand{\shift}{\mathcal{S}}
\newcommand{\fhat}{\hat{f}}
\newcommand{\fdot}{\dot{f}}
\newcommand{\loss}{\mathcal{L}}
\newcommand{\risk}{\mathcal{R}}
\newcommand{\E}{\mathbb{E}}
\newcommand{\N}{\mathcal{N}}
\newcommand{\one}{\mathbf{1}}
\newcommand{\eye}{\mathbf{I}}
\newcommand{\abold}{\bm{\alpha}}
\newcommand{\fbold}{\bm{\varphi}}
\newcommand{\Hbold}{\mathbf{H}}
\newcommand{\hbold}{\mathbf{h}}

\DeclareMathOperator*{\argmin}{arg\,min}
\DeclareMathOperator*{\minimize}{minimize}
\DeclareMathOperator*{\config}{config}
\DeclareMathOperator*{\sgn}{sgn}
\DeclarePairedDelimiter\abs{\lvert}{\rvert}
\DeclarePairedDelimiter\norm{\lVert}{\rVert}

\journal{arXiv.org}

\begin{document}

\begin{frontmatter}

\title{Geostatistical Learning: Challenges and Opportunities\tnoteref{mytitlenote}}
\tnotetext[mytitlenote]{Software is available at \url{https://github.com/IBM/geostats-gen-error}}

\author[impaaddress]{J\'ulio Hoffimann\corref{correspondingauthor}\footnote{%
\textbf{J\'ulio Hoffimann}: Conceptualization, Methodology, Software, Formal Analysis, Investigation, Visualization, Writing - Original Draft %
\textbf{Maciel Zortea}: Methodology, Validation %
\textbf{Breno de Carvalho}: Data Curation, Validation %
\textbf{Bianca Zadrozny}: Methodology, Validation, Supervision%
}}
\cortext[correspondingauthor]{Corresponding author}
\ead{julio.hoffimann@impa.br}

\author[ibmaddress]{Maciel Zortea}
\ead{mazortea@br.ibm.com}

\author[ibmaddress]{Breno de Carvalho}
\ead{brenow@ibm.com}

\author[ibmaddress]{Bianca Zadrozny}
\ead{biancaz@br.ibm.com}

\address[impaaddress]{Instituto de Matem\'atica Pura e Aplicada}
\address[ibmaddress]{IBM Research Brazil}

\begin{abstract}
Statistical learning theory provides the foundation to applied machine learning, and its various successful applications in computer vision, natural language processing and other scientific domains. The theory, however, does not take into account the unique challenges of performing statistical learning in geospatial settings. For instance, it is well known that model errors cannot be assumed to be independent and identically distributed in geospatial (a.k.a. regionalized) variables due to spatial correlation; and trends caused by geophysical processes lead to covariate shifts between the domain where the model was trained and the domain where it will be applied, which in turn harm the use of classical learning methodologies that rely on random samples of the data. In this work, we introduce the \emph{geostatistical (transfer) learning} problem, and illustrate the challenges of learning from geospatial data by assessing widely-used methods for estimating generalization error of learning models, under covariate shift and spatial correlation. Experiments with synthetic Gaussian process data as well as with real data from geophysical surveys in New Zealand indicate that none of the methods are adequate for model selection in a geospatial context. We provide general guidelines regarding the choice of these methods in practice while new methods are being actively researched.
\end{abstract}


\end{frontmatter}


\section{Introduction}\label{sec:intro}

Classical learning theory \citep{Wu1999,Hastie2009,Mitchell1997} and its applied machine learning methods have been popularized in the geosciences after various technological advances, leading initiatives in open-source software \citep{Pedregosa2011,Abadi2016,Paszke2017,Innes2018}, and intense marketing from a diverse portfolio of industries. In spite of its popularity, learning theory cannot be applied straightforwardly to solve problems in the geosciences as the characteristics of these problems violate fundamental assumptions used to derive the theory and related methods (e.g. i.i.d. samples).

Among these methods derived under classical assumptions (more on this later), those for estimating the generalization (or prediction) error of learned models in unseen samples are crucial in practice \citep{Hastie2009}. In fact, estimates of generalization error are widely used for selecting the best performing model for a problem, out of a collection of available models \citep{Arlot2010}. If estimates of error are inaccurate because of violated assumptions, then there is great chance that models will be selected inappropriately \citep{Ferraciolli2019}. The issue is aggravated when models of great expressiveness (i.e. many learning parameters) are considered in the collection since they are quite capable of overfitting the available data \citep{Jiang2019,Zhang2019}. In the following paragraphs, we consider statistical learning broadly as minimization of generalization error.

The literature on generalization error estimation methods is vast \citep{Arlot2010,Vehtari2012}, and we do not intend to review it extensively here. Nevertheless, some methods have gained popularity since their introduction in the mid 70s because of their generality, ease of use, and availability in open-source software:

\paragraph{\textbf{Leave-one-out (1974)}} The leave-one-out method for assessing and selecting learning models was based on the idea that to estimate the prediction error on an unseen sample one only needs to hide a seen sample from a dataset and learn the model. Because the hidden sample has a known label, the method can compare the model prediction with the true label for the sample. By repeating the process over the entire dataset, one gets an estimate of the expected generalization error \citep{Stone1974}. Leave-one-out has been investigated in parallel by many statisticians, including Nicholson (1960) and Stone (1974), and is also known as ordinary cross-validation.

\paragraph{\textbf{k-fold cross-validation (1975)}} The term k-fold cross-validation refers to a family of error estimation methods that split a dataset into non-overlapping ``folds'' for model evaluation. Similar to leave-one-out, each fold is hidden while the model is learned using the remaining folds. It can be thought of as a generalization of leave-one-out where folds may have more than a single sample \citep{Geisser1975,Burman1989}. Cross-validation is less computationally expensive than leave-one-out depending on the size and number of folds, but can introduce bias in the error estimates if the number of samples in the folds used for learning is much smaller than the original number of samples in the dataset.

Major assumptions are involved in the derivation of the estimation methods listed above. The first of them is the assumption that samples come from independent and identically distributed (i.i.d.) random variables. It is well-known that spatial samples are not i.i.d., and that spatial correlation needs to be modeled explicitly with geostatistical theory. Even though the sample mean of the empirical error used in those methods is an unbiased estimator of the prediction error regardless of the i.i.d. assumption, the precision of the estimator can be degraded considerably with non-i.i.d. samples.

Motivated by the necessity to leverage non-i.i.d. samples in practical applications, and evidence that model's performance is affected by spatial correlation \citep{Cracknell2014,Baglaeva2020}, the statistical community devised new error estimation methods using the spatial coordinates of the samples:

\paragraph{\textbf{h-block leave-one-out (1995)}} Developed for time-series data (i.e. data showing temporal dependency), the h-block leave-one-out method is based on the principle that stationary processes achieve a correlation length (the ``h'') after which the samples are not correlated. The time-series data is then split such that samples used for error evaluation are at least "h steps" distant from the samples used to learn the model \citep{Burman1994}. Burman (1994) showed how the method outperformed traditional leave-one-out in time-series prediction by selecting the hyperparameter ``h'' as a fraction of the data, and correcting the error estimates accordingly to avoid bias.

\paragraph{\textbf{Spatial leave-one-out (2014)}} Spatial leave-one-out is a generalization of h-block leave-one-out from time-series to spatial data \citep{LeRest2014}. The principle is the same, except that the blocks have multiple dimensions (e.g. norm-balls).

\paragraph{\textbf{Block cross-validation (2016)}} Similarly to k-fold cross-validation for non-spatial data, block cross-validation was proposed as a faster alternative to spatial leave-one-out. The method creates folds using blocks of size equal to the spatial correlation length, and separates samples for error evaluation from samples used to learn the model. The method introduces the concept of ``dead zones'', which are regions near the evaluation block that are discarded to avoid over-optimistic error estimates \citep{Roberts2017,Pohjankukka2017}.

Unlike the estimation methods proposed in the 70s, which use random splits of the data, these methods split the data based on spatial coordinates and what the authors called ``dead zones''. This set of heuristics for creating data splits avoids configurations in which the model is evaluated on samples that are too near ($<$ spatial correlation length) other samples used for learning the model. Consequently, these estimation methods tend to produce error estimates that are higher on average than their non-spatial counterparts, which are known to be over-optimistic in the presence of spatial correlation. However, systematic splits of the data introduce bias, which have not been emphasized enough in the literature.

All methods for estimating generalization error in classical learning theory, including the methods listed above, rely on a second major assumption. The assumption that the distribution of unseen samples to which the model will be applied is equal to the distribution of samples over which the model was trained. This assumption is very unrealistic for various applications in geosciences, which involve quite heterogeneous (i.e. variable), heteroscedastic (i.e. with different variability) processes \citep{Chiles2012}.

Very recently, an alternative to classical learning theory has been proposed, known as transfer learning theory, to deal with the more difficult problem of learning under shifts in distribution, and learning tasks \citep{Pan2010,Weiss2016,Silver2008}. The theory introduces methods that are more amenable for geoscientific work \citep{Zadrozny2003,Zadrozny2004,Fan2005}, yet these same methods were not derived for geospatial data (e.g. climate data, Earth observation data, field measurements).

Of particular interest in this work, the covariate shift problem is a type of transfer learning problem where the samples on which the model is applied have a distribution of covariates that differs from the distribution of covariates over which the model was trained \citep{Joaquin2008}. It is relevant in geoscientific applications in which a list of explanatory features is known to predict a response via a set of physical laws that hold everywhere. Under covariate shift, a generalization error estimation method has been proposed:

\paragraph{\textbf{Importance-weighted cross-validation (2007)}} Under covariate shift, and assuming that learning models may be misspecified, classical cross-validation is not unbiased. Importance weights can be considered for each sample to recover the unbiasedness property of the method, and this is the core idea of importance-weighted cross-validation \citep{Sugiyama2006,Sugiyama2007}. The method is unbiased under covariate shift for the two most common supervised learning tasks: regression and classification.

The importance weights used in importance-weighted cross-validation are ratios between the target (or test) probability density and the source (or train) probability density of covariates. Density ratios are useful in a much broader set of applications including two-sample tests, outlier detection, and distribution comparison. For that reason, the problem of density ratio estimation has become a general statistical problem \citep{Sugiyama2012}. Various density ratio estimators have been proposed with increasing performance \citep{Huang2007,Sugiyama2009,Kanamori2009,Kanamori2009a}, yet an investigation is missing that contemplates importance-weighted cross-validation and other existing error estimation methods in geospatial settings.

In this work, we introduce \emph{geostatistical (transfer) learning}, and discuss how most prior work in spatial statistics fits in a specific type of learning from geospatial data that we term \emph{pointwise learning}. In order to illustrate the challenges of learning from geospatial data, we assess existing estimators of generalization error from the literature using synthetic Gaussian process data and real data from geophysical well logs in New Zealand that we made publicly available \citep{W.S.RdeCarvalho2020}.

The paper is organized as follows. In \autoref{sec:problem}, we introduce \emph{geostatistical (transfer) learning}, which contains all the elements involved in learning from geospatial data. We define covariate shift in the geospatial setting and briefly review the concept of spatial correlation. In \autoref{sec:error}, we define generalization error in geostatistical learning, discuss how it generalizes the classical definition of error in non-spatial settings, and review estimators of generalization error from the literature devised for \emph{pointwise learning}. In \autoref{sec:exps}, we present our experiments with geospatial data, and discuss the results of different error estimation methods. In \autoref{sec:concls}, we conclude the work and share a few remarks regarding the choice of error estimation methods in practice.

\section{Geostatistical learning}\label{sec:problem}

In this section, we define the elements of statistical learning in geospatial settings. We discuss the covariate shift and spatial correlation properties of the problem, and illustrate how they affect the involved feature spaces.

Consider a sample space $\Omega$, a source spatial domain $\D_s \subset \R^{d_s}$, and a target spatial domain $\D_t \subset \R^{d_t}$ on which stochastic processes (i.e. spatial random variables) are defined:
\begin{equation}
\begin{aligned}
    Z_{s_j}&\colon\,\D_s \times \Omega \to \R,\ j=1,2,\ldots,n_s\ \text{on source domain}\ \D_s\\
    Z_{t_j}&\colon\,\D_t \times \Omega \to \R,\ j=1,2,\ldots,n_t\ \text{on target domain}\ \D_t
\end{aligned}
\end{equation}
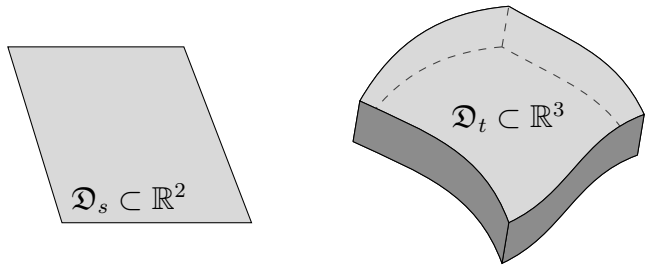
\begin{figure}[h]
\centering
\begin{tikzpicture}[scale=0.18]
\tikzset{thin_line/.style={      thin,               solid, color=black}}
\tikzset{dash_line/.style={      thin,              dashed, color=darkgray}}
\tikzset{vect_line/.style={very thick, ->, >=latex,  solid, color=black}}

\draw [fill=gray!30] plot [mark=none, sharp cycle] coordinates {(7,3) (21,3) (16,16) (3,16)};
\draw (7,3)  node[above right] {\large $\D_s \subset \R^2$};

\coordinate (a)  at ( 40, 3);
\coordinate (am) at (39.5, 0);
\coordinate (an) at ( 44.5, 7);

\coordinate (b)  at ( 50, 11);
\coordinate (bm) at ( 49.5, 8);

\coordinate (c)  at ( 40, 19);
\coordinate (cm) at (39.5, 16);

\coordinate (d)  at (29, 12);
\coordinate (dm) at (28.5, 9);

\draw[thin_line] (a) to[out= 25, in=200] (b);    
\draw[thin_line] (b) to[out=115, in=-30] (c);    
\draw[thin_line] (c) to[out=185, in= 60] (d);    
\draw[thin_line] (d) to[out=-25, in=110] (a);

\draw[fill=gray!30] (a) to[out= 25, in=200] 
                    (b) to[out=115, in=-30] 
                    (c) to[out=185, in= 60] 
                    (d) to[out=-25, in=110] (a);    

\draw[thin_line] (am) to[out= 25, in=200] (bm);    
\draw[dash_line] (bm) to[out=115, in=-30] (cm);    
\draw[dash_line] (cm) to[out=185, in= 60] (dm);    
\draw[thin_line] (dm) to[out=-25, in=110] (am);    

\draw[thin_line] (am) -- (a);
\draw[thin_line] (bm) -- (b);
\draw[dash_line] (cm) -- (c);
\draw[thin_line] (dm) -- (d);
 
\draw[fill=gray!90] (a)  to[out= 25, in=200] 
                    (b)  -- 
                    (bm) to[out=200, in=25]
                    (am) -- (a);

\draw[fill=gray!90] (am) --
                    (a)  to[out=110, in=-25]
                    (d)  --
                    (dm) to[out=-25, in=110] (am);
\draw (a)++(0,6)  node[above] {\large $\D_t \subset \R^3$};
\end{tikzpicture}
\caption{Examples of source and target spatial domains. (a) Source domain $\D_s \subset \R^2$, which may represent a satellite view of an area of interest. (b) Target domain $\D_t \subset \R^3$, which may represent a body of rock in the subsurface of the Earth.}
\label{fig:domains}
\end{figure}

For example, $(Z_{s_j})_{j=1,2,\ldots,n_s}$ may represent a collection of processes observed remotely from satellite on a 2D surface $\D_s \subset \R^2$, whereas $(Z_{t_j})_{j=1,2,\ldots,n_t}$ may represent processes that occur within the 3D subsurface of the Earth $\D_t \subset \R^3$ (see \autoref{fig:domains}). Any process $Z$ in these collections can be viewed in two distinct ways:

\paragraph{\textbf{Geostatistical theory}} From the viewpoint of geostatistical theory, samples $z(\cdot,\omega)$ of the process $Z(\uu,w)$ are obtained by fixing $\omega \in \Omega$. These samples are spatial maps that assign a real number to each location $\uu \in \D$.

\paragraph{\textbf{Learning theory}} From the viewpoint of statistical learning theory, scalar samples $z(\uu,\cdot)$ are obtained by fixing $\uu \in \D$. These scalar samples are ordered into a ``feature vector'' $\x_\uu = (z_1, z_2,\ldots, z_n)$ for a collection of processes $(Z_j)_{j=1,2,\ldots,n}$, and for a specific location $\uu \in \D$. In this case, $\XX_\uu\colon\, \Omega \to \R^n$ denotes the corresponding random vector of features such that $\x_\uu \sim \XX_\uu$.

In order to define the geostatistical learning problem, we need to understand the joint probability distribution of features for all locations in a spatial domain $\Pr(\{\XX_\uu\}_{\uu\in\D})$. This distribution is very complex in general as feature vectors $\XX_\uu$ and $\XX_{\vv}$ for two different locations $\uu \ne \vv$ are not independent. The closer the locations $\uu, \vv \in \D$ in the spatial domain, the more similar are their features $\x_\uu,\,\x_\vv \in \R^n$ in the feature space. Moreover, given that only one realization $z^{obs} = z(\cdot,\omega) \sim Z$ of the process is available at any given time, one must introduce stationarity assumptions inside $\D$ to pool together different scalar samples $z(\uu,\cdot)$ from different locations $\uu \in \D$ in the spatial domain, and be able to estimate the distribution.

Regardless of the stationarity assumptions involved in modeling these processes, we can assume that inside $\D$ the probability $\Pr_\D(\XX) = \Pr(\{\XX_\uu\}_{\uu\in\D})$ is well-defined. For example, most prior art in statistical learning with geospatial data assumes that the pointwise probability of features $\Pr_\uu(\XX) = \Pr(\XX_\uu)$ is not a function of location, that is $\Pr_\uu(\XX) = \Pr(\XX),\, \forall \uu \in \D$. Under this assumption, samples from everywhere in $\D$ are used to estimate $\Pr(\XX) = \Pr(Z_1,Z_2,\ldots,Z_n)$. With the additional assumption that the feature vectors $\XX_\uu$ and $\XX_\vv$ are independent, the joint distribution of features for all locations can be written as $\Pr_\D(\XX) = \prod_{\uu\in\D} \Pr_\uu(\XX)$.

Whereas the pointwise stationarity assumption may be reasonable inside a given spatial domain, the assumption of spatial independence of features is rarely defensible in practice. Additionally, pointwise stationarity often does not transfer from a source domain $\D_s$ where the model is learned to a target domain $\D_t$ where the model is applied, and consequently the joint distributions of features differ $\Pr_{\D_s} \ne \Pr_{\D_t}$. Before we can illustrate these two issues in more detail, we need to complete the definition of geostatistical learning problems by introducing the notion of spatial learning tasks.

We have introduced the notion of spatial domain $\D$, and the notion of joint probability of features $\Pr_\D(\XX)$ for all locations in the domain. Now we introduce the notion of spatial learning tasks, which are similar to classical learning tasks, but with the main difference that they can leverage properties of the underlying spatial domain. Classically, a learning task describes an action in terms of available features to produce new data. For example, ``predict feature $Z_{j_0}$ from features $(Z_{j_1},Z_{j_2})$'', or ``cluster the samples using features $(Z_{j_1},Z_{j_2},Z_{j_3})$'' are classical learning tasks. Differently, a spatial learning task $T$ involves the spatial domain $\D$ besides the features, and is therefore more complex. Practical examples from the industry include:
\begin{itemize}
    \item \textbf{Mining}: The task of segmenting a mineral deposit from borehole samples using a set of features is a spatial learning task. It assumes the segmentation result to be a \emph{contiguous volume} of rock, which is an additional constraint in terms of spatial coordinates.
    \item \textbf{Agriculture}: The task of identifying crops from satellite images is a spatial learning task. Locations that have the same crop type \emph{appear together} in the images despite possible noise in image layers (e.g. presence of clouds, animals).
    \item \textbf{Petroleum}: The task of segmenting formations from seismic data is a spatial learning task because these formations are large-scale \emph{near-horizontal} layers of stacked rock.
\end{itemize}

Many more examples of spatial learning tasks exist, and others are yet to be proposed. Given the concepts introduced above, we are now ready for the main definition of this section:

\paragraph{\textbf{Definition (Geostatistical Learning)}} Let $\D_s$ be a source spatial domain, and $\D_t$ be a target spatial domain. Let $\Pr_{\D_s}(\XX_s)$ and $\Pr_{\D_t}(\XX_t)$ be the joint distributions of features for all locations in these domains, and let $T_s$ and $T_t$ be two spatial learning tasks. Geostatistical (transfer) learning consists of learning $T_t$ over $\D_t$ using the knowledge acquired while learning $T_s$ over $\D_s$, and assuming that the observed spatial data in $\D_s$ and $\D_t$ are both a single spatial sample of $\Pr_{\D_s}(\XX_s)$ and $\Pr_{\D_t}(\XX_t)$, respectively.

There are considerable differences between the classical definition of transfer learning \citep{Pan2010,Weiss2016}, and the proposed definition above. First, the distribution we have denoted by $\Pr_\D(\XX)$ is spatial and involves all the locations $\uu\in\D$, whereas the distribution in classical transfer learning is the marginal for any specific location, obtained from the assumption of pointwise stationarity $\Pr(\XX_\uu) = \Pr(\XX)$. Second, we use the term domain to refer to spatial domains $\D$, whereas the non-spatial literature uses the same term for the pair $\left(\XX_\uu, \Pr(\XX_\uu)\right) = \left(\XX, \Pr(\XX)\right)$. Third, the spatial learning task $T$ we have introduced may be described in terms of properties of the spatial domain, which are not available in generic transfer learning problems.

Having understood the main differences between classical and geostatistical learning, we now focus our attention to a specific type of geostatistical transfer learning problem, and illustrate some of the unique challenges caused by spatial dependence.

\subsection{Covariate shift}\label{sec:shift}

Assume that the two spatial domains are different $\D_s \ne \D_t$, but that they share a set of processes $(Z_1,Z_2,\ldots,Z_n)$. Additionally, assume that pointwise stationarity holds. Let $Z_o = f(Z_1,Z_2,\ldots,Z_n)$ be a new process obtained as a function of the shared processes, and assume that it has only been observed in $\D_s$ via a measuring device and/or manual labeling. That is, $z_o^{obs}(\cdot, \omega) \sim Z_o$ is a spatial sample of the process $Z_o$ over $\D_s$. Under these assumptions, we can introduce the shared vector of features $\XX_s = \XX_t = \XX = (Z_1,Z_2,\ldots,Z_n,Z_o)$, and the supervised learning task $T_s = T_t = T$ of predicting the process $Z_o$ regardless of location $\uu \in \D_s \cup \D_t$.

Let $\X = \XX_{1:n}$ be the explanatory features, and $\Y = \XX_{n+1}$ be the response feature. For any $\uu\in\D_s$, we can write $\Pr_\uu(\X,\Y) = \Pr_\uu(\Y | \X) \Pr_\uu(\X)$. Likewise, for any $\vv\in\D_t$ we can write $\Pr_\vv(\X,\Y) = \Pr_\vv(\Y | \X) \Pr_\vv(\X)$. The covariate shift property is defined as follows:

\paragraph{\textbf{Definition (Covariate Shift)}} A geostatistical learning problem has the covariate shift property when for any $\uu\in\D_s$ and for any $\vv\in\D_t$ the distributions $\Pr_\uu(\X,\Y)$ and $\Pr_\vv(\X,\Y)$ differ by $\Pr_\uu(\X) \ne \Pr_\vv(\X)$ while $\Pr_\uu(\Y | \X) = \Pr_\vv(\Y | \X)$.

The property is based on the idea that the underlying true function $f$ that created the process $\Y = f(\X)$ is the same for all $\uu\in\D_s$ and all $\vv\in\D_t$. In this case, the function is approximated by the conditional distribution $\Pr_\uu(\Y | \X) = \Pr_\vv(\Y | \X)$ for each and every location (see \autoref{fig:shift}).
\tikzset{
  error band/.style={fill=red},
  error band style/.style={
    error band/.append style=#1
  }
}
\newcommand{\addplotwitherrorband}[4][]{
  \addplot [#1, draw=none, stack plots=y, forget plot] {#2-(#3)};
  \addplot +[#1, draw=none, stack plots=y, error band] {(#3)+(#4)} \closedcycle;
  \addplot [#1, draw=none, stack plots=y, forget plot] {-(#2)-(#3)};

  \addplot [#1, forget plot] {#2};
}
\def\bandcolor{gray}
\begin{figure}[h]
\centering
\begin{tikzpicture}[scale=0.75,
declare function={
  f(\x)=rad(\x)-sin(\x);
  s(\x)=1.5+0.5*cos(\x);
}]
\begin{axis}[domain=0:360,
ticks=none,enlargelimits=false,
xlabel={$\X$},ylabel={$\Y$},
extra description/.code={
  \node[rotate=30] at (0.7,0.5) {$\Pr_\uu(\Y | \X) = \Pr_\vv(\Y | \X)$};
  \node at (0.2,0.15) {$\Pr_\uu(\X)$};
  \node at (0.8,0.2) {$\Pr_\vv(\X)$};
  \draw[-Latex] (0.2,0.8) node[inner sep=0,label=above:$f$] {}
    to[out=-90,in=135] (0.5,0.61);
},
cycle list={
  error band style=\bandcolor!20\\
  error band style=\bandcolor!40\\
  error band style=\bandcolor!60\\
  error band style=\bandcolor!80\\
  error band style=\bandcolor!100\\
}]
\foreach \dh in {1,0.5,0.25,0.125,0.0625} {
  \addplotwitherrorband[thick,black] {f(x)}{\dh*s(x)}{\dh*s(x)}
}
\addplot[smooth,fill=gray!80,fill opacity=0.5] coordinates {(0,-5) (180,-3) (270,-5)} --cycle;
\addplot[smooth,fill=purple!80,fill opacity=0.5] coordinates {(180,-5) (270,-3) (360,-5)} --cycle;
\end{axis}
\end{tikzpicture}
\caption{Covariate shift. The true underlying function $\Y = f(\X)$ is approximated equally as $\Pr_\uu(\Y | \X) = \Pr_\vv(\Y | \X),\ \forall\uu\in\D_s,\forall\vv\in\D_t$. The only difference between source and target spatial domains lies in the distributions of explanatory variables $\Pr_\uu(\X) \ne \Pr_\vv(\X)$.}
\label{fig:shift}
\end{figure}
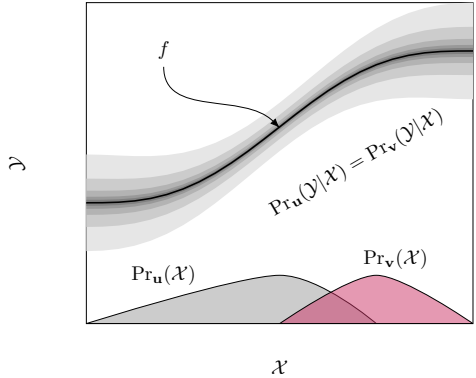

In the geosciences, it is very common to encounter problems with covariate shift due to the great variability of natural processes. Whenever a model is (1) learned using labels provided by experts on a spatial domain $\D_s$, is (2) validated with classical train-validation-test methodologies (meaning that it satisfies some performance threshold), and yet (3) performs poorly on a new spatial domain $\D_t$ where the labeling function is expected to be the same, we can conclude that there are shifts in distribution. In \autoref{sec:exps} we illustrate covariate shifts in real data that we prepared in-house from geophysical surveys in New Zealand.

\subsection{Spatial correlation}\label{sec:corr}

Another important issue with geospatial data that is often ignored is spatial dependence, which we illustrate next. As mentioned earlier, the closer are two locations $\uu,\vv\in\D$ in a spatial domain, the more similar are their features $\x_\uu,\,\x_\vv \in \R^n$ in the feature space. Different statistics are available to quantify this spatial dependence in a collection of samples, and a popular choice from geostatistics is the variogram $\gamma(h)$, which estimates for each spatial lag $h=||\uu-\vv||\in\R^+_0$ a correlation $\sigma^2 - \gamma(h)$, where $\sigma^2$ is the total sill in the samples \citep{Chiles2012}. Parallel algorithms for efficient variogram estimation exist in the literature \citep{Hoffimann2019}, and can be useful tools for fast diagnosis of the spatial correlation property:

\paragraph{\textbf{Definition (Spatial Correlation)}} A geostatistical learning problem has the spatial correlation property when the variogram of any of the stochastic processes $(Z_{s_j})_{j=1,2,\ldots,n_s}$ and $(Z_{t_j})_{j=1,2,\ldots,n_t}$ defined over $\D_s$ and $\D_t$ has a non-negligible positive range (or correlation length).

\begin{figure}[h]
    \centering
    \includegraphics[width=\textwidth]{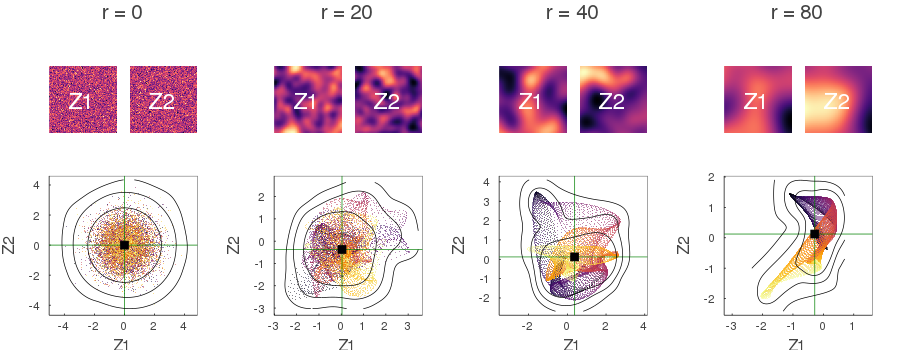}
    \caption{Impact of spatial correlation in feature space. Two Gaussian processes $Z_1$ and $Z_2$ are simulated over a domain $\D$ with $100\times 100$ pixels. As the variogram range $r$ increases from $r=0$ to $r=80$ pixels, the distribution of features $\Pr(\X) = \Pr(Z_1,Z_2)$ is gradually deformed away from the standard Gaussian $\N(0,I)$. Marker colors in scatter plots represent the pixel number in column-major order.}
    \label{fig:feat-corr}
\end{figure}

Besides serving as a tool for diagnosing spatial correlation in geostatistical learning problems, variograms can also be used to simulate spatial processes with theoretical correlation structure. In \autoref{fig:feat-corr}, we illustrate the impact of spatial correlation in the feature space of two independent spatial processes $Z_1$ and $Z_2$ simulated with direct (a.k.a. LU) Gaussian simulation \citep{Alabert1987}. As we increase the variogram range $r$ in a spatial domain $\D$ with $100\times 100$ pixels, we observe that the distribution of features $\Pr(\X) = \Pr(Z_1, Z_2)$ is gradually deformed from a standard Gaussian ($r=0$) to a ``boomerang''-shaped distribution ($r=80$).

\begin{figure}[h]
    \centering
    \includegraphics[width=\textwidth]{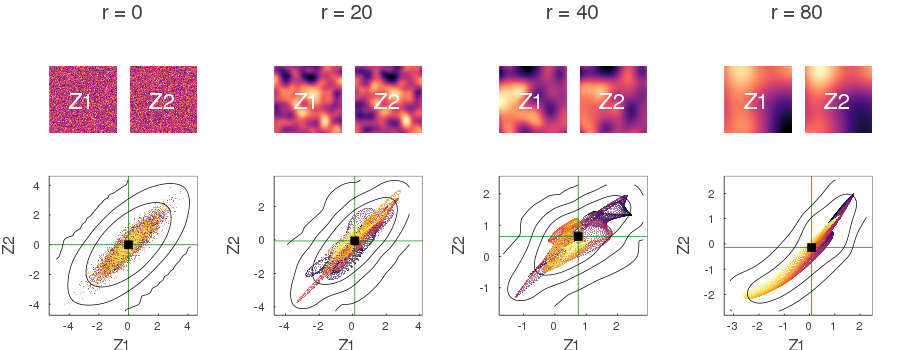}
    \caption{Impact of spatial correlation in feature space with correlated processes. Similar to \autoref{fig:feat-corr} with the difference that the processes $Z_1$ and $Z_2$ are strongly correlated with correlation $\rho(Z_1,Z_2) = 0.9$.}
    \label{fig:feat-corr2}
\end{figure}

Similar deformations are observed when the two processes $Z_1$ and $Z_2$ are correlated. In \autoref{fig:feat-corr2}, we illustrate the impact of spatial correlation for an inter-process correlation of $\rho(Z_1,Z_2) = 0.9$.

Spatial correlation may have different impact in source and target domains $\D_s$ and $\D_t$, and can certainly affect the generalization error of learning models. In our experiments of \autoref{sec:exps}, we assume that the variogram range of source and target processes are equal (i.e. $r_s = r_t = r$) to facilitate the analysis of the results. In practice, source and target processes may also have different spatial correlation, which is a type of shift that is not considered in classical transfer learning problems.

\section{Generalization error of learning models}\label{sec:error}

Having defined geostatistical learning problems, and their covariate shift and spatial correlation properties, we now turn into a general definition of generalization error of learning models in geospatial settings. We review an importance-weighted approximation of a related generalization error based on pointwise stationarity assumptions, and the use of an efficient importance-weighted cross-validation method for error estimation.

Consider a geostatistical learning problem $\problem = \left\{(\D_s,\Pr_{\D_s},T_s), (\D_t,\Pr_{\D_t},T_t)\right\}$ with a single supervised spatial learning task $T_s = T_t = T$ (e.g. regression), and assume that a set of response features $\Y_\uu$ are created by a function $f$, based on a set of explanatory features $\X_\uu$ for each and every location $\uu\in \D_s \cup \D_t$. Our goal is to learn a model $\{\Y_\uu\}_{\uu\in\D_t} \approx \fhat\Big(\{\X\}_{\uu\in\D_t}\Big)$ over the target domain $\D_t$ that approximates $f$ in terms of expected risk for some spatial supervised loss function $\loss$:
\begin{equation}\label{eq:risk}
\fhat = \argmin_g \E_{\Pr_{\D_t}}\left[\loss\Bigg(\{\Y_\uu\}_{\uu\in\D_t},\ g\Big(\{\X\}_{\uu\in\D_t}\Big)\Bigg)\right]
\end{equation}

In the expected value of \autoref{eq:risk}, spatial samples of the processes are drawn from $\Pr_{\D_t}$ and rearranged into feature vectors $\X_\uu$ and $\Y_\uu$ for every location $\uu \in \D_t$. The spatial loss function $\loss$ compares the spatial map of features from the sample $\{\Y_\uu\}_{\uu\in\D_t}$ with the approximated map from the model $g\Big(\{\X\}_{\uu\in\D_t}\Big)$. The model $\fhat$ is the model that minimizes the expected loss (or risk) over the target domain $\D_t$.

\paragraph{\textbf{Definition (Generalization Error)}} The generalization error of a learning model $\fhat$ in a geostatistical learning problem $\problem$ is the expected risk attained by the model when spatial samples are drawn from $\Pr_{\D_t}$ over the target domain $\D_t$ (see \autoref{eq:risk}).

Unlike the classical definition of generalization error, the definition above for geostatistical learning problems relies on a spatial loss function $\loss$, and on spatial samples like those produced via geostatistical simulation \citep{Hoffimann2017,Mariethoz2010}. For truly spatial learning models $\fhat$ that use multiple locations in the spatial domain to make predictions, this generalization error is more appropriate. In this present work, however; we do not target spatial learning models, and only consider pointwise learning:

\paragraph{\textbf{Definition (Pointwise Learning)}} Given a family of classical (non-spatial) learning models $\{\fhat_\uu\}_{\uu\in\D}$, pointwise learning consists of learning the model $\fhat\Big(\{\X\}_{\uu\in\D}\Big) = \{\fhat_\uu(\X_\uu)\}_{\uu\in\D}$ that assigns for each location $\uu\in\D$ the value $\fhat_\uu(\X_\uu)$ independently of the explanatory features at other locations.

More specifically, we consider pointwise learning with families that are made of a single learning model $\{\fhat_\uu\}_{\uu\in\D} = \{\fdot\}$. In this case, the model $\fdot$ is often learned based on pointwise stationarity assumptions, for some pointwise loss $\dot{\loss}$:
\begin{equation}\label{eq:pointrisk}
\fdot = \argmin_{\dot{g}} \E_{\Pr}\left[\dot{\loss}\Big(\Y, \dot{g}(\X)\Big)\right]
\end{equation}

Although pointwise learning with a single model is a very simple type of geostatistical learning, it is by far the most widely used approach in the geospatial literature. We acknowledge this fact, and consider an empirical approximation of the pointwise expected risk in \autoref{eq:pointrisk} as opposed to the spatial expected risk in \autoref{eq:risk}.

An empirical approximation of the pointwise expected risk of a model $\dot{g}$ can be obtained via discretization of the target spatial domain $\D_t$:
\begin{equation}\label{eq:emprisk}
\risk_t(\dot{g}) = \E_{\Pr_{\uu\in\D_t}}\left[\dot{\loss}\Big(\Y, \dot{g}(\X)\Big)\right] \approx \frac{1}{|\D_t|} \sum_{\uu\in\D_t} \dot{\loss}\Big(\Y_\uu, \dot{g}(\X_\uu)\Big)
\end{equation}
with $|\D_t|$ the number of locations in the discretization. The problem with this empirical approximation is that the response features $\Y_\uu$ are not available in the target domain where the model will be applied. However, it is easy to show that the pointwise expected risk in \autoref{eq:emprisk} can be rewritten with importance weights $\dot{w}(\X,\Y)$ when samples from $\D_s$ are drawn instead \citep{Sugiyama2006}:
\begin{equation}
\risk_t(\dot{g}) = \E_{\Pr_{\vv\in\D_t}}\left[\dot{\loss}\Big(\Y, \dot{g}(\X)\Big)\right] = \E_{\Pr_{\uu\in\D_s}}\left[\dot{w}(\X,\Y)\dot{\loss}\Big(\Y, \dot{g}(\X)\Big)\right]
\end{equation}
with $\dot{w}(\X,\Y) = \frac{\Pr_{\vv\in\D_t}(\X,\Y)} {\Pr_{\uu\in\D_s}(\X,\Y)}$. Under covariate shift, the importance weights only depend on the distribution of explanatory features $\dot{w}(\X) = \frac{\Pr_{\vv\in\D_t}(\X)} {\Pr_{\uu\in\D_s}(\X)}$, and we can write a simple importance-weighted empirical approximation:
\begin{equation}\label{eq:iwer}
\risk_t(\dot{g}) = \E_{\Pr_{\uu\in\D_s}}\left[\dot{w}(\X)\dot{\loss}\Big(\Y, \dot{g}(\X)\Big)\right] \approx \frac{1}{|\D_s|} \sum_{\uu\in\D_s} \dot{w}(\X_\uu)\dot{\loss}\Big(\Y_\uu, \dot{g}(\X_\uu)\Big)
\end{equation}

Our goal is to find the pointwise model that minimizes the empirical risk approximation $\hat{\risk}_t(\dot{g})$ introduced in \autoref{eq:iwer}:
\begin{equation}
\fdot = \argmin_{\dot{g}} \hat{\risk}_t(\dot{g})
\end{equation}

Alternatively, our goal is to rank a collection of models $\{\dot{g}_i\}_{i=1,2,\ldots,k}$ based on their empirical risk $\{\hat{\risk}_t(\dot{g}_i)\}_{i=1,2,\ldots,k}$ in a geostatistical learning problem to aid model selection.

In order to achieve the stated goals, we need to (1) estimate the importance weights in the empirical risk approximation, and (2) remove the dependence of the approximation on a specific dataset. These two issues are addressed in the following sections.

\subsection{Density ratio estimation}\label{sec:dre}

The empirical approximation of the risk $\hat{\risk}_t(\dot{g})$, depends on estimates of the weights $\dot{w}(\X_\uu)$, which are ratios of probabilities in the target and source domains. The following problem can be posed \citep{Sugiyama2012}:

\paragraph{\textbf{Definition (Density Ratio Estimation)}} Given two collections of samples $\{\X_\uu\}_{\uu\in\D_s}$ and $\{\X_\vv\}_{\vv\in\D_t}$ from source and target domains, estimate the density ratio $\frac{\Pr_{\vv\in\D_t}(\X)}{\Pr_{\uu\in\D_s}(\X)}$ at any new sample $\X$. In particular, estimate the ratio at all samples $\{\X_\uu\}_{\uu\in\D_s}$ from the source.

Efficient methods for density ratio estimation that perform well with high-dimensional features have been proposed in the literature. In this work we consider a fast method named Least Squares Importance Fitting (LSIF) \citep{Kanamori2009,Kanamori2009a}. The LSIF method assumes that the weights are a linear combination of basis functions $\dot{w}(\X_\uu) = \abold^\top \fbold(\X_\uu)$ with coefficients to be learned $\abold = (\alpha_1,\alpha_2,\ldots,\alpha_b)$ and fixed basis $\fbold(\X_\uu) = (\varphi_1(\X_\uu),\varphi_2(\X_\uu),\ldots,\varphi_b(\X_\uu))$. The LSIF estimator is derived by minimizing the squared error with the true density ratio:
\begin{equation}
\begin{aligned}
\minimize_{\abold\in\R^b} \quad & \frac{1}{2}\int\left(\dot{w}(\X_\uu) - \frac{\Pr_{\vv\in\D_t}(\X_\uu)}{\Pr_{\uu\in\D_s}(\X_\uu)} \right)^2\Pr_{\uu\in\D_s}(\X_\uu) d\X_\uu\\
\textrm{s.t.} \quad & \abold \succeq \mathbf{0}
\end{aligned}
\end{equation}
under the constraint that densities are always positive. By choosing $b$ center features randomly from the target domain $\{\X_i\}_{i=1,2,\ldots,b}$, the method introduces a Gaussian kernel basis $\varphi_i(\X_\uu) = k(\X_\uu,\X_i)$ that simplifies the objective function to a matrix form:
\begin{equation}
\begin{aligned}
\minimize_{\abold\in\R^b} \quad & \frac{1}{2} \abold^\top \Hbold \abold - \hbold^\top \abold + \lambda \mathbf{1}^\top \abold\\
\textrm{s.t.} \quad & \abold \succeq \mathbf{0}
\end{aligned}
\end{equation}
with $\Hbold = \int \fbold(\X_\uu) \fbold(\X_\uu)^\top \Pr_{\uu\in\D_s}(\X_\uu) d\X_\uu$ and $\hbold = \int \fbold(\X_\vv) \Pr_{\vv\in\D_t}(\X_\vv)d\X_\vv$. The regularization parameter $\lambda \ge 0$ on the coefficients $\abold$ avoids overfitting, and empirical estimates of both $\Hbold$ and $\hbold$ are easily obtained with sample averages:
\begin{equation}
\begin{aligned}
\hat{\Hbold} &= \frac{1}{|\D_s|}\sum_{\uu\in\D_s} \fbold(\X_\uu)\fbold(\X_\uu)^\top\\
\hat{\hbold} &= \frac{1}{|\D_t|}\sum_{\vv\in\D_t} \fbold(\X_\vv)
\end{aligned}
\end{equation}

This quadratic optimization problem with linear inequality constraints can be solved very efficiently with modern optimization software \citep{KMogensen2018,Dunning2017}. In the end, the optimal coefficients $\abold^\star$ are plugged back into the basis expansion for optimal estimates of the weights on new samples $\dot{w}(\X) = {\abold^\star}^\top \fbold(\X)$.

\subsection{Weighted cross-validation}\label{sec:iwcv}

In order to remove the dependence of the empirical risk approximation on the dataset, we use importance-weighted cross-validation (IWCV) \citep{Sugiyama2006,Sugiyama2007}. As with the traditional cross-validation procedure, the source domain is split into $k$ folds $\D_s = \bigcup_{j=1}^k \D_s^{(j)}$, and each fold $\D_s^{(j)}$ is hidden for error evaluation while the model $\dot{g}^{(j)}$ is learned on the remaining folds:
\begin{equation}
\hat{\risk}_t^{IWCV}(\dot{g}) = \frac{1}{k} \sum_{j=1}^k \frac{1}{|\D_s^{(j)}|} \sum_{\uu\in\D_s^{(j)}} (\dot{w}(\X_\uu))^l \dot{\loss}\Big(\Y_\uu, \dot{g}^{(j)}(\X_\uu)\Big)
\end{equation}

The main difference in the IWCV procedure are the weights that multiply each sample. The regularization exponent $l \in [0,1]$ can be set to zero to recover the traditional estimator, or to a positive value to account for covariate shift. An optimal value for $l$ can be found via hyperparameter search by considering another layer of cross-validation. In this work, we simply set default values for $l$ such as $l=1$ or $l=0.5$.

In the rest of the paper, we combine IWCV with LSIF into a method for estimating generalization error that we term \emph{Density Ratio Validation}. Although IWCV is known to outperform classical cross-validation methods in non-spatial settings, little is known about its performance with geospatial data. Moreover, like all prior art, IWCV approximates the pointwise generalization error of \autoref{eq:pointrisk} as opposed to the geostatistical generalization error of \autoref{eq:risk}, and therefore is limited by design to non-spatial learning models.

\section{Experiments}\label{sec:exps}

In this section, we perform experiments to assess estimators of generalization error under varying covariate shifts and spatial correlation lengths. We consider Cross-Validation (CV), Block Cross-Validation (BCV) and Density Ratio Validation (DRV), which all rely on the same cross-validatory mechanism of splitting data into folds.

First, we use synthetic Gaussian process data and simple labeling functions to construct geostatistical learning problems for which learning models have a known (via geostatistical simulation) generalization error. In this case, we assess the estimators in terms of how well they estimate the actual error under various spatial distributions. Second, we demonstrate how the estimators are used for model selection in a real application with well logs from New Zealand, which can be considered to be a dataset of moderate size in this field.

\subsection{Gaussian processes}\label{sec:gauss}

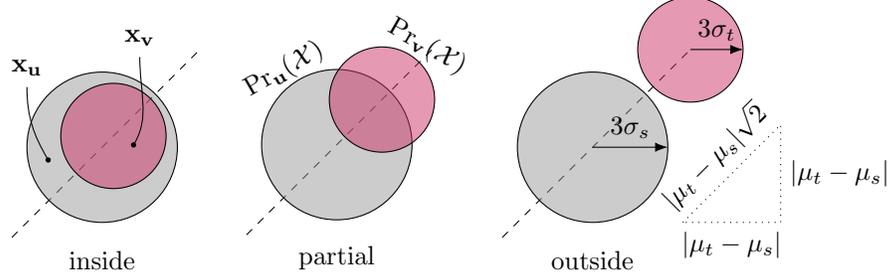
\begin{figure}[h]
\centering%
\begin{subfigure}[b]{.3\textwidth}
\begin{tikzpicture}
\draw[dashed] (-1.2,-1.2) -- (1.3,1.3) {};
\draw[fill=gray!80,fill opacity=0.5] (0,0) circle (1);
\draw[fill=purple!80,fill opacity=0.5] (0.15,0.15) circle (0.7);
\node at (0,-1.5) {inside};
\draw[-{Circle[length=2pt]}] (-1.0,0.8) node[inner sep=0,label=above:$\x_\uu$] {} to[out=-90,in=120] (-0.7,-0.2);
\draw[-{Circle[length=2pt]}] (0.5,1.2) node[inner sep=0,label=above:$\x_\vv$] {} to[out=-90,in=60] (0.4,0.0);
\node at (0,2.2) {For $\uu\in\D_s$ and $\vv\in\D_t$:};
\end{tikzpicture}
\end{subfigure}%
\begin{subfigure}[b]{.3\textwidth}
\begin{tikzpicture}
\draw[dashed] (-1.2,-1.2) -- (1.3,1.3) {};
\draw[fill=gray!80,fill opacity=0.5] (0,0) circle (1);
\draw[fill=purple!80,fill opacity=0.5] (0.6,0.6) circle (0.7);
\node at (0,-1.5) {partial};
\node[rotate=30] at (-0.7,1.0) {$\Pr_\uu(\X)$};
\node[rotate=-30] at (1.2,1.3) {$\Pr_\vv(\X)$};
\end{tikzpicture}
\end{subfigure}%
\begin{subfigure}[b]{.3\textwidth}
\begin{tikzpicture}
\draw[dashed] (-1.2,-1.2) -- (1.3,1.3) {};
\draw[fill=gray!80,fill opacity=0.5] (0,0) circle (1);
\draw[fill=purple!80,fill opacity=0.5] (1.3,1.3) circle (0.7);
\node at (0,-1.5) {outside};
\draw[-Latex] (0,0) -- (1,0) node[midway,above] {$3\sigma_s$};
\draw[-Latex] (1.3,1.3) -- (1.3+0.7,1.3) node[midway,above] {$3\sigma_t$};
\draw[dotted] (0+1.2,0-1) -- (1.3+1.2,1.3-1) node[midway,above,rotate=45] {$\abs{\mu_t - \mu_s}\sqrt{2}$} -- (1.3+1.2,0-1) node[midway,right] {$\abs{\mu_t - \mu_s}$} -- cycle node[midway,below] {$\abs{\mu_t - \mu_s}$};
\end{tikzpicture}
\end{subfigure}%
\caption{Three possible shift configurations. The target distribution is ``inside'' the source distribution (left), ``outside'' (right), or they show ``partial'' overlap (middle). The processes $Z_{s_1}$ and $Z_{s_2}$ are observed at any location $\uu\in\D_s$ forming a feature vector $\x_\uu\in\R^2$. Similarly, the processes $Z_{t_1}$ and $Z_{t_2}$ (which are shifted versions of the previous two) are observed at any location $\vv\in\D_t$ forming a feature vector $\x_\vv\in\R^2$. In this case the features $\x_\uu \sim \Pr_\uu(\X)$ and $\x_\vv \sim \Pr_\vv(\X)$ are samples of bivariate Gaussians illustrated with circles of radii $3\sigma_s$ and $3\sigma_t$ centered at $\mu_s\one$ and $\mu_t\one$.}
\label{fig:config}
\end{figure}

Let $Z_{s_1},Z_{s_2}$ be two Gaussian processes with constant mean $\mu_s$ and variogram $\gamma_s$ defined over $\D_s$, and likewise let $Z_{t_1},Z_{t_2}$ be two Gaussian processes with constant mean $\mu_t$ and variogram $\gamma_t$ defined over $\D_t$. Denote by $r_s$ the variogram range (or correlation length) and by $\sigma_s^2$ the variogram sill (or total variance) of the processes in the source domain. Likewise, denote by $r_t$ and $\sigma_t^2$ the range and sill of the variogram in the target domain. It is clear that pointwise stationarity holds inside each of these domains. The feature vector $\X_\uu \in \R^2$ for any location $\uu\in\D_s$ in the source domain has a bivariate Gaussian distribution $\N(\mu_s \one, \sigma_s^2 \eye)$, whereas the feature vector $\X_\vv \in \R^2$ for any location $\vv\in\D_t$ in the target domain has a bivariate Gaussian distribution $\N(\mu_t \one, \sigma_t^2 \eye)$. By constraining the variogram ranges to be equal in source and target domains, that is $r_s = r_t = r$, and by requiring that both variograms pass through the origin (i.e. no nugget effect), we can investigate two types of covariate shift for various ranges $r$:

\paragraph{\textbf{Mean shift}} Define the shift in the mean as $\delta = c\, \abs{\mu_t - \mu_s} \in [0,\infty)$ for some normalization constant $c > 0$. In this experiment, we set the value of this constant to $c = \frac{1}{3\sqrt{2}\sigma_s}$ for convenience so that a $\delta=1$ becomes equivalent to $\abs{\mu_t - \mu_s} = 3\sqrt{2}\sigma_s$, which in turn is equivalent to two circles (i.e. bivariate Gaussians) of radii $3\sigma_s$ touching each other along the identity line, see \autoref{fig:config}.

\paragraph{\textbf{Variance shift}} Define the shift in the variance as $\tau = \sigma_t / \sigma_s \in (0,\infty)$. Here, $\tau=1$ means absence of variance shift, $\tau < 1$ means that the variance where the model is applied is smaller than the variance where the model was trained, and $\tau > 1$ means the exact opposite of $\tau < 1$.

Geostatistical learning problems with $\tau>1$ are very challenging to solve, and usually require additional extrapolation models, beyond the pointwise learning models discussed in this work. Therefore, we only consider cases with $\tau \le 1$ in this experiment. More specifically, we consider all combinations of shift in the mean and variance of Gaussian features by varying $(\delta,\tau)$ in the open unit square $\B = [0,1]\times(0,1]$.

Given a shift parameterized by $(\delta,\tau) \in \B$, we can classify it into one of three possible configurations depending on how the source and target distributions of features overlap:
\begin{equation}\label{eq:config}
\config(\delta,\tau) =
\begin{cases}
\text{inside}, & 2\delta \le 1 - \tau\\
\text{outside}, & 2\delta \ge 1 + \tau\\
\text{partial}, & \text{otherwise}
\end{cases}
\end{equation}

The first configuration in \autoref{eq:config} refers to the case in which the target distribution $\N(\mu_t \one, \sigma_t^2 \eye)$ is ``inside'' the source distribution $\N(\mu_s \one, \sigma_s^2 \eye)$ meaning that the circle of radius $3\sigma_t$ centered at $\mu_t\one$ is contained in the circle of radius $3\sigma_s$ centered at $\mu_s\one$. Similarly, the second configuration refers to the case in which the target distribution is ``outside'' the source distribution. Finally, the third configuration refers to a ``partial'' overlap when the two distributions share a common set of samples but are not entirely one inside of the other. We note, however; that the illustration with circles provided in \autoref{fig:config} is only representative in the absence of spatial correlation (i.e. $r = 0$), see \autoref{fig:feat-corr}.

To efficiently simulate multiple spatial samples of the processes over a regular grid domain with $100\times 100$ locations (or pixels), we use spectral Gaussian simulation \citep{Gutjahr1997}. We fix the parameters of the source distribution at $\mu_s = 0$ and $\sigma_s = 1$ without loss of generality, and assume no inter-process correlation (i.e. $\rho = 0$) like we did in \autoref{fig:feat-corr}. Under these modeling assumptions, we are able to investigate the spatial distribution of features as a function of shift parameters $(\delta,\tau) \in \B$ and variogram ranges $r \in \C = \{0,10,20\}$.

\begin{figure}[h]
\centering
\begin{tikzpicture}
\begin{groupplot}[group style={group size=3 by 1, horizontal sep=2cm},
                  height=4cm,width=4cm,
                  xlabel=$Z_1$,ylabel=$Z_2$,
                  ylabel near ticks,
                  enlargelimits=false]
    \nextgroupplot[title={$p=1,w=2$}]
        \addplot graphics[xmin=-4,ymin=-4,xmax=4,ymax=4] {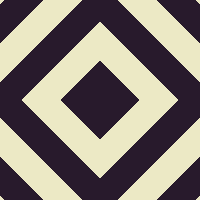};
    \nextgroupplot[title={$p=1,w=4$}]
        \addplot graphics[xmin=-4,ymin=-4,xmax=4,ymax=4] {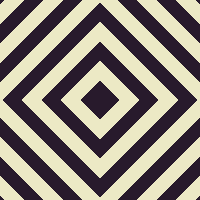};
    \nextgroupplot[title={$p=2,w=4$}]
        \addplot graphics[xmin=-4,ymin=-4,xmax=4,ymax=4] {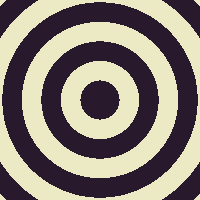};
\end{groupplot}
\end{tikzpicture}
\caption{Labeling function $f_{p,w}(\x) = \sgn(\sin(w\norm{\x}_p))$. Labels form alternating patterns in the feature space for different $p$-norm and angular frequencies $w$.}
\label{fig:label}
\end{figure}

To fully specify the geostatistical learning problem, we need to specify a learning task. The task consists of predicting a binary variable $\Y_\vv$ at locations $\vv$ in the target grid $\D_t$ based on observations $y_\uu$ of the variable at locations $\uu$ in the source grid $\D_s$. These observations (or labels) are synthesized using known labeling functions such as $y_\uu = \sgn(\sin(w\norm{\x_\uu}_p))$, where $\norm{\cdot}_p$ is the $p$-norm, $w$ is the angular frequency, and $\sgn$ is the modified sign function that assigns $+1$ to $x\ge0$ and $-1$ otherwise. The observations produced by these functions form alternating patterns in the feature space, which are not trivial to predict with simple learning models, see \autoref{fig:label}. In this experiment, we fix $p=1$ and $w = 4$ to save computational time. Other norms and angular frequencies produce similar results.

Having defined the problem, we proceed and specify learning models in order to investigate the different estimators of generalization error. We choose two models that are based on different prediction mechanisms \citep{Hastie2009}:

\paragraph{\textbf{Decision tree}} A pointwise decision tree model $\dot{f}_T$ makes predictions solely based on the features of the sample, without exploiting nearby features in the feature space.

\paragraph{\textbf{K-nearest neighbors}} A pointwise k-nearest neighbors model $\dot{f}_N$ makes predictions based on nearby features, and is sometimes called a ``spatial model''.

These two models $\dot{f}_T$ and $\dot{f}_N$ are simply representative models from the ``non-spatial'' and ``spatial'' families of models. We emphasise, however; that the term ``spatial model'' can be misleading in the spatial statistics literature. It is important to distinguish ``spatial models'' such as k-nearest neighbors that exploit the notion of proximity of features in the \emph{feature space} from ``geospatial models'' that also exploit the proximity of samples in the \emph{physical space} (or spatial domain as we have been calling it) besides their features.

The experiment proceeds as follows. For each shift $(\delta,\tau) \in \B$, each correlation length $r \in \C$, and each pointwise learning model $\dot{f} \in \{\dot{f}_T, \dot{f}_N\}$, we sample a problem $\problem_{\delta,\tau,r}$ and estimate the generalization error of the model $\dot{f}$ on the problem with CV, BCV and DRV. We set the hyperparameters of the CV and BCV estimators based on the fact that the correlation length never exceeds $20$. For instance, we set the block side in BCV to $s=20$, and use the equivalent number of folds in CV, i.e. $k=(100/20)^2=25$ for a domain with $100\times 100$ pixels. We set the kernel width of the LSIF estimator in DRV to $\sigma=2$ based on the synthetic Gaussian distributions, and use $10$ kernels in the basis expansion. Additionally, we approximate the true generalization error of the models with Monte Carlo simulation over the target domain (e.g. 100 spatial samples).

To facilitate the visualization of the results, we introduce shift functions $\shift\colon \B \to [0,\infty)$ that map the shift parameters $(\delta,\tau)$ to a single covariate shift value, which can be interpreted loosely as the ``difficulty'' of the problem:

\paragraph{\textbf{Kullback-Leibler divergence}} The divergence or relative entropy between two distributions $p$ and $q$ is defined as $\shift_{KL}(p||q) = \int p(x) \log\frac{p(x)}{q(x)}dx$, and can be derived analytically for two (2D) Gaussian distributions $p = \N(\mu_t \one, \sigma_t^2 \eye)$ and $q = \N(\mu_s \one, \sigma_s^2 \eye)$. We derive a formula in terms of $\delta$ and $\tau$ by fixing $\sigma_s = 1$:
\begin{equation}
    \shift_{KL}(\delta,\tau) = \delta^2 + \tau^2 - \log(\tau^4) - 1
\end{equation}

\paragraph{\textbf{Jaccard distance}} The Jaccard index between two sets $A$ and $B$ is defined as $J(A,B) = \frac{|A\cap B|}{|A\cup B|}$, and the corresponding distance as $\shift_J(A,B) = 1 - J(A,B)$. For two (2D) Gaussian distributions, we consider $A$ and $B$ to be circles of radii $3\sigma_s$ and $3\sigma_t$ centered at $\mu_s\one$ and $\mu_t\one$. The distance is then expressed in terms of areas, which can be derived analytically in terms of $\delta$ and $\tau$ by fixing $\sigma_s = 1$:
\begin{equation}\label{eq:jaccard}
\begin{aligned}
|A| = 9\pi,\quad |B| = 9\pi\tau^2\\
C_1 = 9\arccos\left(\frac{2\delta^2 + 9(1-\tau^2)}{6\sqrt{2}\delta}\right)\\
C_2 = 9\tau^2\arccos\left(\frac{2\delta^2 + 9(\tau^2 - 1)}{6\sqrt{2}\delta\tau}\right)\\
C_3 = \frac{1}{2}\sqrt{\left(9(1+\tau)^2 - 2\delta^2\right)\left(2\delta^2 - 9(1-\tau)^2\right)}\\
|A \cap B| = C_1 + C_2 - C_3\\
|A \cup B| = |A| + |B| - |A \cap B|
\end{aligned}
\end{equation}

\paragraph{\textbf{Novelty factor}} We propose a new shift function termed the \emph{novelty factor} inspired by the geometric view of Jaccard. First, we define the novelty of $B$ with respect to $A$ as $N(B/A) = \frac{|B- A\cap B| - |A\cap B|}{|B|}$, and notice that it is the fraction of $B$ that is outside of $A$ minus the fraction of $B$ that is inside of $A$. Second, we restrict the definition to cases with $|B| \le |A|$ (e.g. Gaussian case with $\tau \le 1$), and notice that the novelty $N(B/A)$ lies in the interval $[-1,1]$. Finally, we define the novelty factor $\shift_N(A,B) = \frac{N(B/A) + 1}{2}$ in the interval $[0,1]$, which can be easily computed for the Gaussian case using the formulas derived in \autoref{eq:jaccard}.

\begin{figure}[h]
\centering
\includegraphics[width=\textwidth]{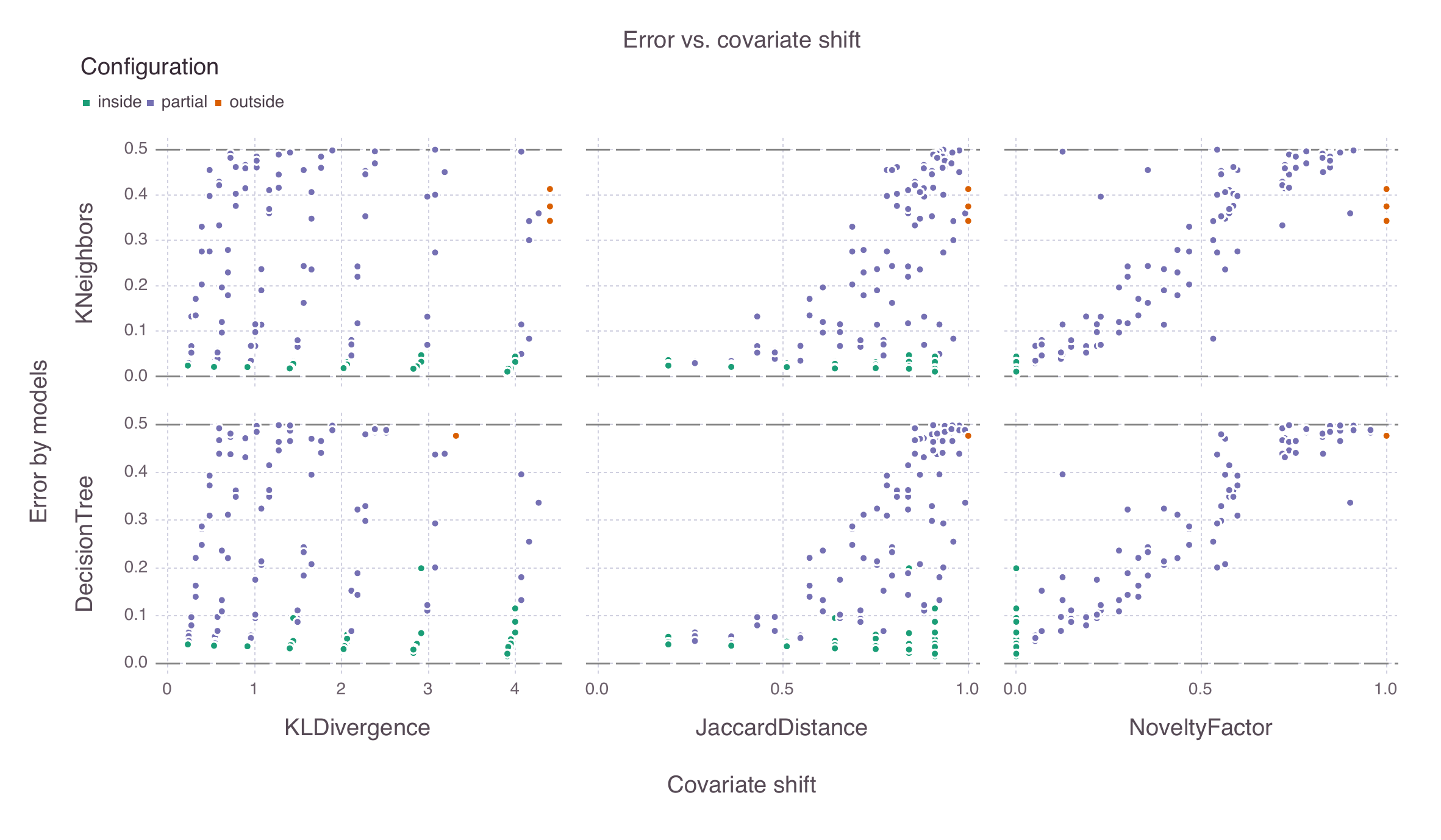}
\caption{Generalization error of learning models versus covariate shift functions. Among all shift functions, the novelty factor is the only function that groups shift configurations along the horizontal axis. Models behave similarly in terms of generalization error for the given dataset size ($100\times 100$ pixels), and difficult shift configurations lead to errors that approach the theoretical Bernoulli limit of $0.5$.}
\label{fig:gaussian-plot1}
\end{figure}

We plot the true generalization error of the models as a function of the different covariate shifts in \autoref{fig:gaussian-plot1}, and color the points according to their shift configuration (see \autoref{eq:config}). In this plot, the horizontal dashed line intercepting the vertical axis at $0.5$ represents a model that assigns positive and negative labels to samples at random with equal weight (i.e. Bernoulli variable).

Among the three shift functions, the novelty factor is the only function that groups shift configurations along the horizontal axis. In this case, configurations deemed easy (i.e. where the target distribution is \emph{inside} the source distribution) appear first, then configurations of moderate difficulty (i.e. \emph{partial} overlap) appear next, and finally difficult configurations (i.e. target is \emph{outside} the source) appear near the theoretical Bernoulli limit. The Kullback-Leibler divergence and the Jaccard distance fail to summarize the shift parameters into a one-dimensional visualization, and are therefore omitted in the next plots.

The two models behave similarly in terms of generalization error for the given dataset size (i.e. $100\times 100$ pixels), and therefore can be aggregated into a single scatter to increase the confidence in the observed trends.

\begin{figure}[h]
\centering
\includegraphics[width=\textwidth]{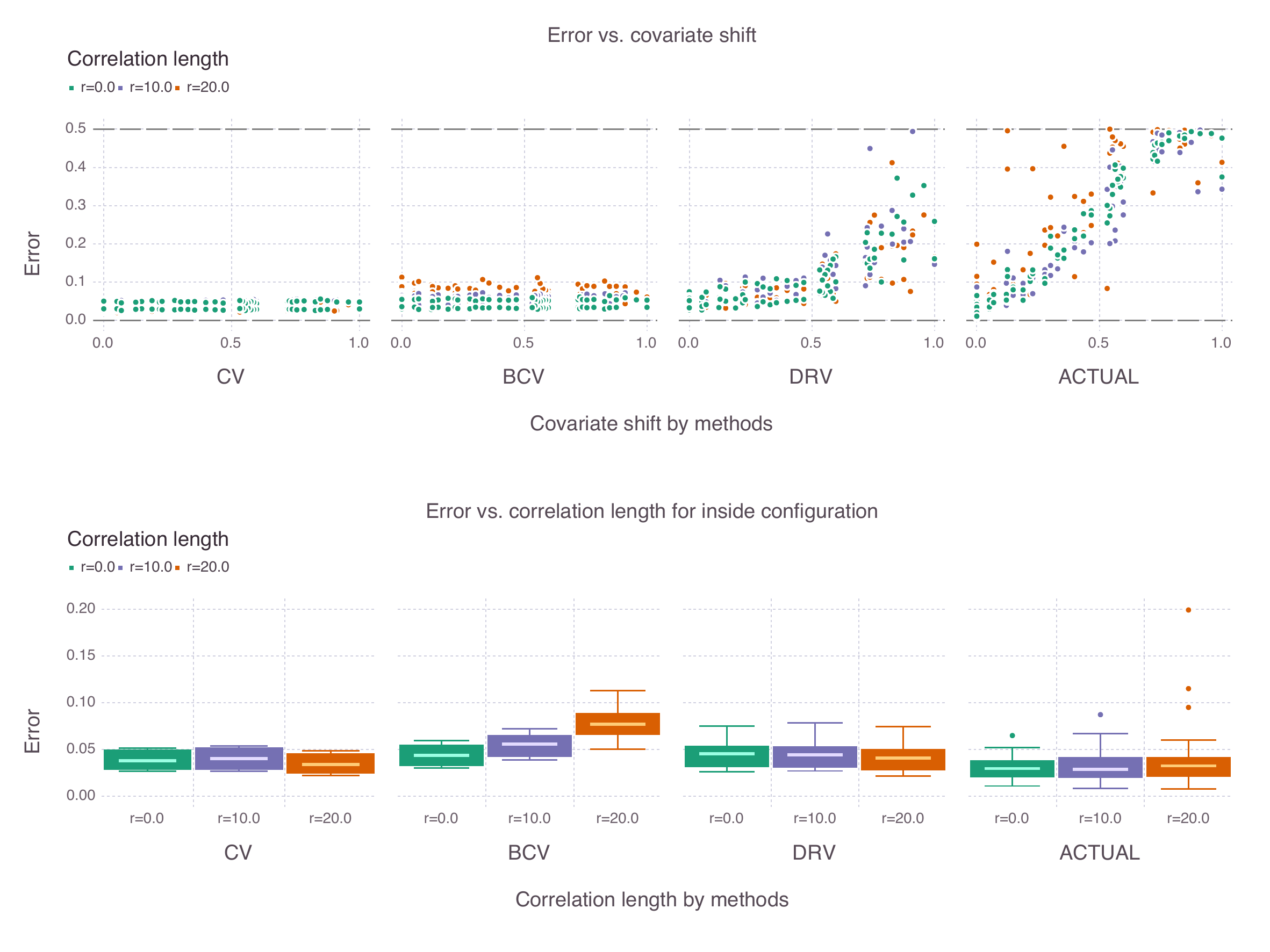}
\caption{Estimates of generalization error for various shifts (i.e. novelty factor) and various correlations lengths. The box plots for the \emph{inside} configuration illustrate how the estimators behave differently for increasing correlation lengths.}
\label{fig:gaussian-plot2}
\end{figure}

We plot the CV, BCV and DRV estimates of generalization error versus covariate shift (i.e. novelty factor) in the top row of \autoref{fig:gaussian-plot2}, and color the points according to their correlation length. We omit a few DRV estimates that suffered from numerical instability in challenging \emph{partial} or \emph{outside} configurations, and show the box plot of error estimates in the bottom row of the figure.

First, we emphasize that the CV and BCV estimates remain constant as a function of covariate shift. This is expected given that these estimators do not make use of the target distribution. The DRV estimates increase with covariate shift as expected, but do not follow the same rate of increase of the true (or actual) generalization error obtained with Monte Carlo simulation. Second, we emphasize in the box plots for the \emph{inside} configuration that the correlation length affects the estimators differently. The CV estimator becomes more optimistic with increasing correlation length, whereas the BCV estimator becomes less optimistic, a result that is also expected from prior art. Additionally, the interquartile range of the BCV estimator increases with correlation length. It is not clear from the box plots that a trend exists for the DRV estimator. The actual generalization error behaves erratically in the presence of large correlation lengths as indicated by the scatter and box plots.

\begin{figure}[h]
\centering
\includegraphics[width=\textwidth]{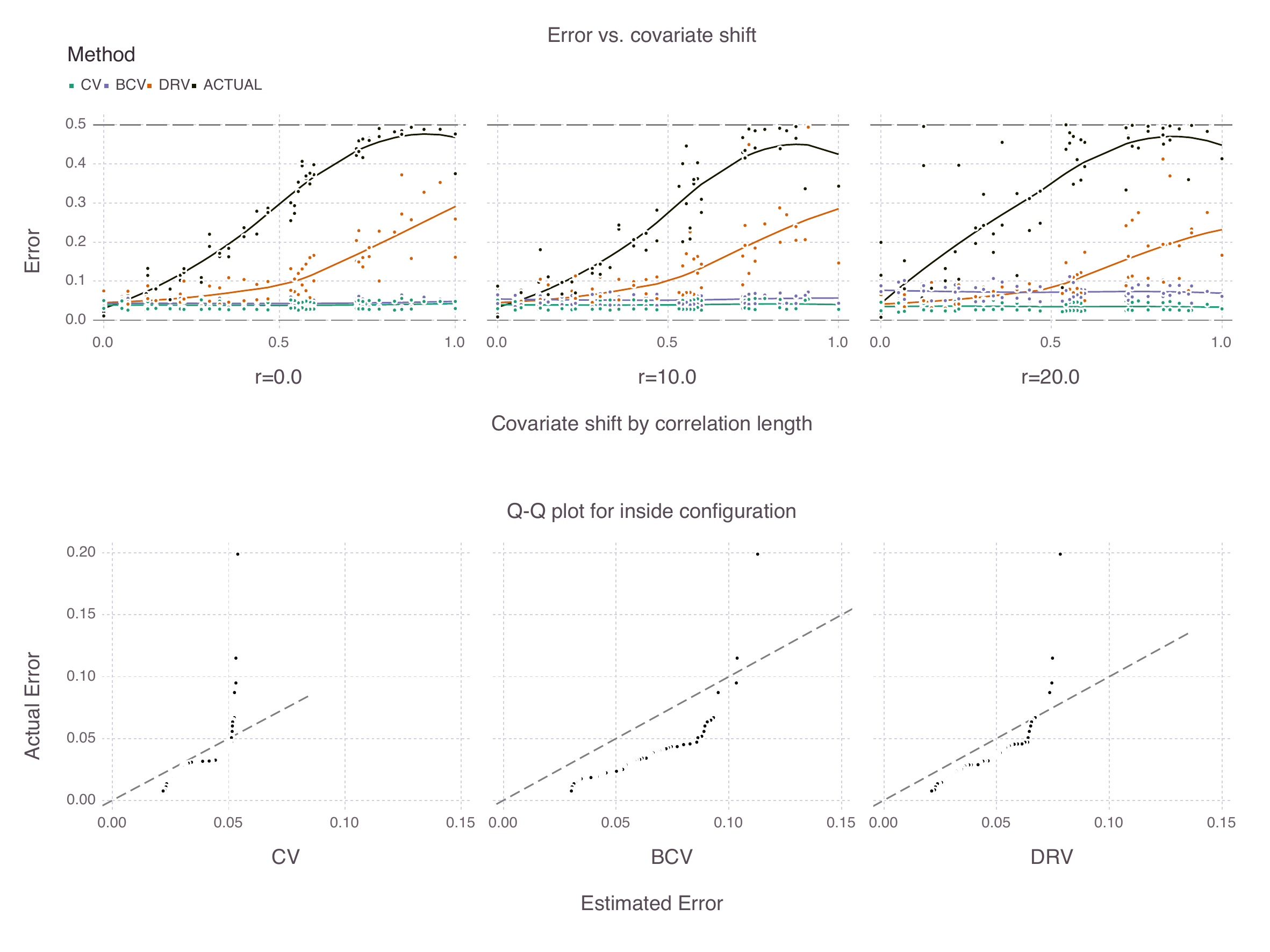}
\caption{Trends of generalization error for different estimators (top) and Q-Q plots of estimated versus actual error for \emph{inside} configuration (bottom).}
\label{fig:gaussian-plot3}
\end{figure}

In order to better visualize the trends in the estimates, we smooth the scatter plots with locally weighted regression per correlation length in the top row of \autoref{fig:gaussian-plot3}, and show in the bottom row of the figure the Q-Q plots of the different estimates against the actual generalization error for the \emph{inside} configuration where all estimators are supposed to perform well.

From the figure, there exists a gap between the DRV estimates and the actual generalization error of the models for all covariate shifts. This gap is expected given that the target distribution may be very different from the source distribution, particularly in \emph{partial} or \emph{outside} shift configurations. On the other hand, the gap also seems to be affected by the correlation length, and is largest with $20$ pixels of correlation. Additionally, we emphasize in the Q-Q plots that the BCV estimates are biased due to the systematic selection of folds. The BCV estimates are less optimistic than the CV estimates, which is a desired property in practice, however there is no guarantee that the former estimates will approximate well the actual generalization error of the models.

\subsection{New Zealand dataset}\label{sec:newzealand}

Unlike the previous experiment with synthetic Gaussian process data and known generalization error, this experiment consists of applying the CV, BCV and DRV estimators to a real dataset of well logs prepared in-house \citep{W.S.RdeCarvalho2020}. We quickly describe the dataset, introduce the related geostatistical learning problems, and use error estimates to rank learning models. Finally, we compare these ranks with an ideal rank obtained with additional label information that is not available during the learning process.

\begin{figure}[h]
\centering
\includegraphics[width=0.5\textwidth]{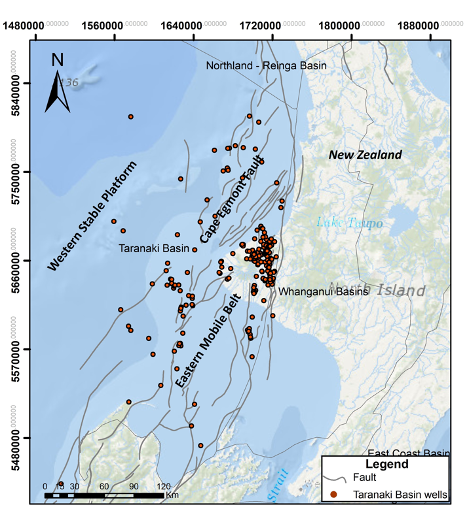}
\caption{Curated dataset with $407$ wells in the Taranaki basin, New Zealand. The basin comprises an area of about $330.000km^2$, located bradly onshore and offshore the New Zealand west coast.}
\label{fig:newzealand}
\end{figure}

The dataset consists of $407$ wells in the Taranaki basin, including the main geophysical logs and reported geological formations. The basin comprises an area of about $330.000km^2$, located broadly onshore and offshore the New Zealand west coast (see \autoref{fig:newzealand}). Well trajectories are georeferenced in UTM coordinates (X and Y) and true vertical depth (Z).

\begin{figure}[h]
\centering
\includegraphics[width=\textwidth]{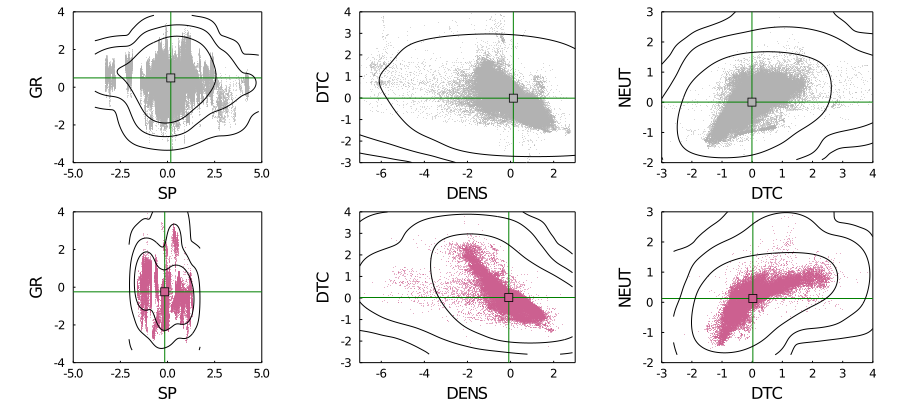}
\caption{Distribution of main geophysical logs onshore (gray) and offshore (purple) centered by the mean and divided by the standard deviation. Visible covariate shift in the scatter and contour plots.}
\label{fig:scatter}
\end{figure}

We split the wells into onshore and offshore locations in order to introduce a geostatistical learning problem with covariate shift. The problem consists of predicting the rock formation from well logs offshore after learning a model with well logs and reported (i.e. manually labeled) formations onshore. The well logs considered are gamma ray (GR), spontaneous potential (SP), density (DENS), compressional sonic (DTC) and neutron porosity (NEUT). We eliminate locations with missing values for these logs and investigate a balanced dataset with the two most frequent formations---Urenui and Manganui. We normalize the logs and illustrate the covariate shift property by comparing the scatter plots of onshore and offshore locations in \autoref{fig:scatter}. Additionally, we define a second geostatistical learning problem without covariate shift. In this case, we join all locations filtered in the previous problem and sample two new sets of locations with sizes respecting the same source-to-target proportion (e.g. $300000 : 50000$).

We set the hyperparameters of the error estimators based on variography and according to available computational resources. In particular, we set blocks for the BCV estimators with sides $10000\times 10000\times 500$ that are much greater than the vertical and horizontal correlation lengths estimated from empirical variograms. We obtain the corresponding number of folds $k=99$ for the CV estimator by partitioning the bounding box of onshore wells into blocks with the given sides. Similarly to the previous experiment with synthetic Gaussian process data, we set the kernel width in DRV to $\sigma=2$ given that the well logs were normalized to have unit variance. Finally, we select a list of learning models to rank including Ridge classification (Ridge), logistic regression (Logistic), k-nearest neighbors (KNeighbors), naive Bayes (GaussianNB), linear discriminant analysis (BayesianLDA), perceptron (Perceptron), decision tree (DecisionTree), and a dummy model that reproduces the marginal distribution of formations in the source domain (Dummy).

In \autoref{tab:onoff}, we report the results for the onshore-to-offshore problem. In the upper part of the table we compare side-by-side the error estimates of the different methods. We highlight the closest estimates to the target error in {\color{NavyBlue} \textbf{blue}} color, and the most distant in {\color{Maroon} \textbf{red}} color. We emphasize that the target error is the error of the model in one single realization of the process, and is \emph{not} the generalization error averaged over multiple spatial realizations. In spite of this important distinction, we still think it is valuable to compare it with the estimates of generalization error given by CV, BCV and DRV since these methods were all derived under pointwise learning assumptions, and are therefore smooth averages over multiple points exactly like the error estimated from the single realization of the target. In the bottom part of the table, we report the model ranks derived from the error estimates as well as the ideal rank derived from the target error.
\begin{table}[h]
\centering
\begin{tabular}{c|cc|ccc}
\hline
\textbf{MODEL} & \textbf{SOURCE} & \textbf{TARGET} & \textbf{CV} & \textbf{BCV} & \textbf{DRV} \\\hline
Ridge & 0.262 & \textbf{0.291} & 0.263 & \color{Maroon}{\textbf{0.326}} & \color{NavyBlue}{\textbf{0.278}} \\
Logistic & 0.262 & \textbf{0.294} & \color{Maroon}{\textbf{0.262}} & 0.322 & \color{NavyBlue}{\textbf{0.275}} \\
KNeighbors & 0.023 & \textbf{0.355} & \color{Maroon}{\textbf{0.034}} & \color{NavyBlue}{\textbf{0.227}} & 0.043 \\
GaussianNB & 0.256 & \textbf{0.321} & \color{Maroon}{\textbf{0.256}} & \color{NavyBlue}{\textbf{0.288}} & 0.284 \\
BayesianLDA & 0.262 & \textbf{0.291} & 0.261 & \color{Maroon}{\textbf{0.326}} & \color{NavyBlue}{\textbf{0.294}} \\
Perceptron & 0.375 & \textbf{0.395} & \color{Maroon}{\textbf{0.344}} & 0.369 & \color{NavyBlue}{\textbf{0.392}} \\
DecisionTree & 0.0 & \textbf{0.378} & \color{Maroon}{\textbf{0.034}} & \color{NavyBlue}{\textbf{0.249}} & 0.043 \\
Dummy & 0.499 & \textbf{0.497} & \color{NavyBlue}{\textbf{0.499}} & 0.502 & \color{Maroon}{\textbf{0.524}} \\\hline
\end{tabular}

\begin{tabular}{cccc}
\hline
\textbf{TARGET RANK} & \textbf{CV RANK} & \textbf{BCV RANK} & \textbf{DRV RANK} \\\hline
\cellcolor[RGB]{221,221,139}{\color{white}{\textbf{BayesianLDA}}} & \cellcolor[RGB]{125,143,145}{\color{white}{\textbf{KNeighbors}}} & \cellcolor[RGB]{125,143,145}{\color{white}{\textbf{KNeighbors}}} & \cellcolor[RGB]{125,143,145}{\color{white}{\textbf{KNeighbors}}} \\
\cellcolor[RGB]{195,195,133}{\color{white}{\textbf{Ridge}}} & \cellcolor[RGB]{83,119,133}{\color{white}{\textbf{DecisionTree}}} & \cellcolor[RGB]{83,119,133}{\color{white}{\textbf{DecisionTree}}} & \cellcolor[RGB]{83,119,133}{\color{white}{\textbf{DecisionTree}}} \\
\cellcolor[RGB]{181,181,145}{\color{white}{\textbf{Logistic}}} & \cellcolor[RGB]{161,166,152}{\color{white}{\textbf{GaussianNB}}} & \cellcolor[RGB]{161,166,152}{\color{white}{\textbf{GaussianNB}}} & \cellcolor[RGB]{181,181,145}{\color{white}{\textbf{Logistic}}} \\
\cellcolor[RGB]{161,166,152}{\color{white}{\textbf{GaussianNB}}} & \cellcolor[RGB]{221,221,139}{\color{white}{\textbf{BayesianLDA}}} & \cellcolor[RGB]{181,181,145}{\color{white}{\textbf{Logistic}}} & \cellcolor[RGB]{195,195,133}{\color{white}{\textbf{Ridge}}} \\
\cellcolor[RGB]{125,143,145}{\color{white}{\textbf{KNeighbors}}} & \cellcolor[RGB]{181,181,145}{\color{white}{\textbf{Logistic}}} & \cellcolor[RGB]{195,195,133}{\color{white}{\textbf{Ridge}}} & \cellcolor[RGB]{161,166,152}{\color{white}{\textbf{GaussianNB}}} \\
\cellcolor[RGB]{83,119,133}{\color{white}{\textbf{DecisionTree}}} & \cellcolor[RGB]{195,195,133}{\color{white}{\textbf{Ridge}}} & \cellcolor[RGB]{221,221,139}{\color{white}{\textbf{BayesianLDA}}} & \cellcolor[RGB]{221,221,139}{\color{white}{\textbf{BayesianLDA}}} \\
\cellcolor[RGB]{45,100,131}{\color{white}{\textbf{Perceptron}}} & \cellcolor[RGB]{45,100,131}{\color{white}{\textbf{Perceptron}}} & \cellcolor[RGB]{45,100,131}{\color{white}{\textbf{Perceptron}}} & \cellcolor[RGB]{45,100,131}{\color{white}{\textbf{Perceptron}}} \\
\cellcolor[RGB]{5,89,140}{\color{white}{\textbf{Dummy}}} & \cellcolor[RGB]{5,89,140}{\color{white}{\textbf{Dummy}}} & \cellcolor[RGB]{5,89,140}{\color{white}{\textbf{Dummy}}} & \cellcolor[RGB]{5,89,140}{\color{white}{\textbf{Dummy}}} \\\hline
\end{tabular}

\caption{Estimates of generalization error with different estimators for the onshore-to-offshore problem. The CV estimator produces estimates that are the most distant to the actual target error due to covariate shift and spatial correlation. None of the estimators is capable of ranking the models correctly. They all select complex models with low generalization ability.}
\label{tab:onoff}
\end{table}

Among the three estimators of generalization error, the CV estimator produces estimates that are the most distant from the target error, with a tendency to underestimate the error. The BCV estimator produces estimates that are higher than the CV estimates, and consequently closer to the target error in this case. The DRV estimator produces the closest estimates for most models, however; like the CV estimator it fails to approximate the error for models like KNeighbors and DecisionTree that are over-fitted to the source distribution. The three estimators fail to rank the models under covariate shift and spatial correlation. Over-fitted models with low generalization ability are incorrectly ranked at the top of the list, and the best models, which are simple ``linear'' models, appear at the bottom. We compare these results with the results obtained for the problem without covariate shift in \autoref{tab:noshift}.
\begin{table}[h]
\centering
\begin{tabular}{c|cc|ccc}
\hline
\textbf{MODEL} & \textbf{SOURCE} & \textbf{TARGET} & \textbf{CV} & \textbf{BCV} & \textbf{DRV} \\\hline
Ridge & 0.267 & \textbf{0.273} & \color{NavyBlue}{\textbf{0.269}} & \color{Maroon}{\textbf{0.329}} & 0.259 \\
Logistic & 0.266 & \textbf{0.271} & \color{NavyBlue}{\textbf{0.265}} & \color{Maroon}{\textbf{0.324}} & 0.254 \\
KNeighbors & 0.026 & \textbf{0.04} & \color{NavyBlue}{\textbf{0.04}} & \color{Maroon}{\textbf{0.232}} & 0.042 \\
GaussianNB & 0.265 & \textbf{0.268} & \color{NavyBlue}{\textbf{0.265}} & \color{Maroon}{\textbf{0.308}} & 0.262 \\
BayesianLDA & 0.267 & \textbf{0.273} & \color{NavyBlue}{\textbf{0.266}} & \color{Maroon}{\textbf{0.329}} & 0.264 \\
Perceptron & 0.445 & \textbf{0.442} & \color{Maroon}{\textbf{0.362}} & \color{NavyBlue}{\textbf{0.402}} & 0.392 \\
DecisionTree & 0.0 & \textbf{0.039} & \color{NavyBlue}{\textbf{0.039}} & \color{Maroon}{\textbf{0.244}} & 0.041 \\
Dummy & 0.499 & \textbf{0.498} & 0.499 & \color{NavyBlue}{\textbf{0.497}} & \color{Maroon}{\textbf{0.476}} \\\hline
\end{tabular}

\begin{tabular}{cccc}
\hline
\textbf{TARGET RANK} & \textbf{CV RANK} & \textbf{BCV RANK} & \textbf{DRV RANK} \\\hline
\cellcolor[RGB]{221,221,139}{\color{white}{\textbf{DecisionTree}}} & \cellcolor[RGB]{221,221,139}{\color{white}{\textbf{DecisionTree}}} & \cellcolor[RGB]{195,195,133}{\color{white}{\textbf{KNeighbors}}} & \cellcolor[RGB]{221,221,139}{\color{white}{\textbf{DecisionTree}}} \\
\cellcolor[RGB]{195,195,133}{\color{white}{\textbf{KNeighbors}}} & \cellcolor[RGB]{195,195,133}{\color{white}{\textbf{KNeighbors}}} & \cellcolor[RGB]{221,221,139}{\color{white}{\textbf{DecisionTree}}} & \cellcolor[RGB]{195,195,133}{\color{white}{\textbf{KNeighbors}}} \\
\cellcolor[RGB]{181,181,145}{\color{white}{\textbf{GaussianNB}}} & \cellcolor[RGB]{181,181,145}{\color{white}{\textbf{GaussianNB}}} & \cellcolor[RGB]{181,181,145}{\color{white}{\textbf{GaussianNB}}} & \cellcolor[RGB]{161,166,152}{\color{white}{\textbf{Logistic}}} \\
\cellcolor[RGB]{161,166,152}{\color{white}{\textbf{Logistic}}} & \cellcolor[RGB]{161,166,152}{\color{white}{\textbf{Logistic}}} & \cellcolor[RGB]{161,166,152}{\color{white}{\textbf{Logistic}}} & \cellcolor[RGB]{125,143,145}{\color{white}{\textbf{Ridge}}} \\
\cellcolor[RGB]{125,143,145}{\color{white}{\textbf{Ridge}}} & \cellcolor[RGB]{83,119,133}{\color{white}{\textbf{BayesianLDA}}} & \cellcolor[RGB]{83,119,133}{\color{white}{\textbf{BayesianLDA}}} & \cellcolor[RGB]{181,181,145}{\color{white}{\textbf{GaussianNB}}} \\
\cellcolor[RGB]{83,119,133}{\color{white}{\textbf{BayesianLDA}}} & \cellcolor[RGB]{125,143,145}{\color{white}{\textbf{Ridge}}} & \cellcolor[RGB]{125,143,145}{\color{white}{\textbf{Ridge}}} & \cellcolor[RGB]{83,119,133}{\color{white}{\textbf{BayesianLDA}}} \\
\cellcolor[RGB]{45,100,131}{\color{white}{\textbf{Perceptron}}} & \cellcolor[RGB]{45,100,131}{\color{white}{\textbf{Perceptron}}} & \cellcolor[RGB]{45,100,131}{\color{white}{\textbf{Perceptron}}} & \cellcolor[RGB]{45,100,131}{\color{white}{\textbf{Perceptron}}} \\
\cellcolor[RGB]{5,89,140}{\color{white}{\textbf{Dummy}}} & \cellcolor[RGB]{5,89,140}{\color{white}{\textbf{Dummy}}} & \cellcolor[RGB]{5,89,140}{\color{white}{\textbf{Dummy}}} & \cellcolor[RGB]{5,89,140}{\color{white}{\textbf{Dummy}}} \\\hline
\end{tabular}

\caption{Estimates of generalization error with different estimators for the problem without covariate shift. The BCV estimator produces estimates that are the most distant to the actual target error due to bias from its systematic selection of folds. All estimators are capable of ranking the models in the absence of covariate shift.}
\label{tab:noshift}
\end{table}

From \autoref{tab:noshift}, the CV estimator produces estimates that are the closest to the target error. The BCV estimator produces estimates that are higher than the CV estimates as before, however this time this means that the BCV estimates are the most distant to the target error. The DRV estimator produces estimates that are not the closest nor the most distant to the target error. The three estimators successfully rank the models from simple linear models at the bottom of the list to more complex learning models at the top. Unlike the previous problem with covariate shift, this time complex models like KNeighbors and DecisionTree show high generalization ability.

\section{Conclusions}\label{sec:concls}

In this work, we introduce \emph{geostatistical (transfer) learning}, and demonstrate how most prior art in statistical learning with geospatial data fits into a category we term \emph{pointwise learning}. We define geostatistical generalization error and demonstrate how existing estimators from the spatial statistics literature such as block cross-validation are derived for that specific category of learning, and are therefore unable to account for general spatial errors.

We propose experiments with spatial data to compare estimators of generalization error, and illustrate how these estimators fail to rank models under covariate shift and spatial correlation. Based on the results of these experiments, we share a few remarks related to the choice of estimators in practice:
\begin{itemize}
    \item The apparent quality of the BCV estimator is falsified in the Q-Q plots of \autoref{fig:gaussian-plot3} and in \autoref{tab:noshift}. The systematic bias produced by the blocking mechanism only guarantees that the error estimates are higher than the CV estimates. When the CV estimates are good (i.e. no covariate shift), the BCV estimates are unnecessarily pessimistic.
    \item The CV estimator is not adequate for geostatistical learning problems that show various forms of covariate shift. Situations without covariate shift are rare in geoscientific settings, and since the DRV estimator works reasonably well for both situations (i.e. with and without shift), it is recommended instead.
    \item Nevertheless, both the CV and DRV estimators suffer from a serious issue with over-fitted models in which case they largely underestimate the generalization error. For risk-averse applications where one needs to be careful about the generalization error of the model, the BCV estimator can provide more conservative results.
    \item None of the three estimators were capable of ranking models correctly under covariate shift and spatial correlation. This is an indication that one needs to be skeptical about interpreting similar rankings available in the literature.
\end{itemize}

Finally, we believe that this work can motivate methodological advances in learning from geospatial data, including research on new estimators of geostatistical generalization error as opposed to pointwise generalization error, and more explicit treatments of spatial coordinates of samples in learning models.

\section*{Computer code availability}

All concepts and methods developed in this paper are made available in the GeoStats.jl project \citep{Hoffimann2018}. The project is hosted on GitHub under the ISC\footnote{\url{https://opensource.org/licenses/ISC}} open-source license: \url{https://github.com/JuliaEarth/GeoStats.jl}.

Experiments of this specific work can be reproduced with the following scripts: \url{https://github.com/IBM/geostats-gen-error}.

\section*{References}

\bibliography{bibliography}

\end{document}